\documentclass[11pt]{article}

\usepackage[margin=1in]{geometry}
\usepackage{amsmath, amssymb}
\usepackage{graphicx}
\usepackage{hyperref}
\usepackage{booktabs}
\usepackage{algorithm}
\usepackage{algpseudocode}
\usepackage{enumitem}
\usepackage{xcolor}

\hypersetup{
    colorlinks=true,
    linkcolor=blue!50!black,
    citecolor=blue!50!black,
    urlcolor=blue!50!black
}

\title{A Mimetic Detector for Adversarial Image Perturbations}

\author{%
   Johnny Corbino\\
   \small Lawrence Berkeley National Laboratory\\
   \small Berkeley, CA, USA\\
   \small \texttt{jcorbino@lbl.gov}
}

\date{\today}

\begin{document}

\maketitle

\begin{abstract}
Adversarial attacks fool deep image classifiers by adding tiny, almost
invisible noise patterns to a clean image.  The standard
$\ell^\infty$-bounded attacks (FGSM and PGD) produce high-frequency,
near-random sign patterns at the pixel level: small in $\ell^2$, but carrying disproportionate
gradient energy.  We exploit this with a single-shot,
training-free detector using the high-order Corbino--Castillo mimetic
operators from the open-source MOLE library.  No retraining, no surrogate classifier, no access to the network
under attack: the verdict is a property of the input alone, computed in
$O(HW)$ time.  We illustrate the detector on the standard \texttt{peppers}
test image: untargeted FGSM and PGD attacks at the
$\ell^\infty$ budget $\varepsilon = 16/255$ flip SqueezeNet's prediction
from \emph{bell pepper} to \emph{doormat} (FGSM) and \emph{maraca} (PGD),
and the detector separates these adversarial inputs from the clean image
by $4.1\times$--$5.0\times$ (FGSM) and $1.9\times$--$2.2\times$ (PGD).
The margin grows monotonically with the operator order $k$, while an
equal-amplitude smooth perturbation leaves the statistic within $1\%$ of
its clean value.
\end{abstract}

\section{The problem and the idea}

\paragraph{Adversarial perturbations.}
Given a trained image classifier $f$ and a clean image $x$, an
\emph{adversarial perturbation} is a small noise pattern $\delta$ such that $f$
misclassifies $x + \delta$ even though $x$ and $x + \delta$ look identical
to a human.  Goodfellow et al.\ \cite{Goodfellow2015} gave the simplest
recipe:
\[
   \delta \;=\; \varepsilon \,\operatorname{sign}\!\bigl(\nabla_x \mathcal{L}(f, x, y)\bigr),
\]
the \emph{Fast Gradient Sign Method} (FGSM); throughout we use the budget
$\varepsilon = 16/255$ at 8-bit pixel depth, a standard choice for
$\ell^\infty$ attacks on ImageNet-scale classifiers.  The
$\ell^\infty$-bounded follow-up attacks (PGD \cite{Madry2018} and, when
run at a fixed budget rather than minimizing it, the $\ell^\infty$
variant of Carlini--Wagner \cite{Carlini2017}) sharpen the FGSM-style
attack and inherit its shape: pixel-wise, $|\delta_{ij}| \le \varepsilon$, with
many entries at or near $\pm \varepsilon$ and little long-range spatial
correlation.  Visually: a uniform high-frequency noise pattern.

\paragraph{Detection vs.\ defense.}
Making $f$ \emph{robust} to such $\delta$ is a hard problem.  A weaker but
useful goal, and the one we pursue here, is \emph{detection}: given an
input $x'$, decide whether it has been perturbed.  The closest existing detector in spirit is SpectralDefense
\cite{Harder2021}, which trains a classifier on Fourier-domain
features of the input; ours is training-free and returns a single scalar.

\paragraph{The angle.}
A clean natural image has gradient energy concentrated on its edges, with
mostly smooth content elsewhere.  An adversarial perturbation, by contrast,
has gradient energy spread \emph{uniformly} across the entire image at the
highest possible spatial frequency.  Comparing the gradient energy of an
input to its plain pixel norm gives a single number that is small for clean
images and large for adversarial ones.

\paragraph{Why mimetic operators?}
On a finite pixel grid, ``gradient'' needs a discretization.  We use the
\emph{Corbino--Castillo mimetic gradient operators} \cite{Corbino2020JCAM},
implemented in the open-source MOLE library \cite{MOLE}, because they:
\begin{itemize}[leftmargin=*,itemsep=2pt]
   \item Provide high-order accuracy with no free parameters.
   \item Use the \emph{same} order of accuracy at the boundary as in the
         interior, so that rectangular framing does not artificially
         trip the detector.
   \item Come with a companion diagonal, positive-definite weight matrix
         $P$ of the same order $k$, reducing the gradient-energy
         computation to a single weighted inner product.
   \item Are obtained in two lines of MATLAB/Octave:\
         \texttt{G = grad2D($k$, $W$, $1$, $H$, $1$)} and
         \texttt{P = weightsP2D($k$, $W$, $1$, $H$, $1$)}.
\end{itemize}

\section{The detector}

We define the \emph{mimetic gradient-energy ratio}
\[
   T(x) \;=\; \frac{(G\,u)^{\!\top}\, P\,(G\,u)}{\sum_{i,j} x_{ij}^2},
\]
where $u$ is the image $x$ padded with one ghost cell on each side
(replicate boundary) and vectorized in MOLE's $x$-fastest order, $G$ is
the order-$k$ Corbino--Castillo mimetic gradient, and $P$ is the diagonal,
positive-definite weight matrix associated with $G$.  The numerator is a
discrete gradient energy (sum of squared gradient values weighted by the
diagonal entries of $P$); the denominator is the plain Euclidean sum of
squared pixel values.  $T$ is dimensionless,
scale-invariant ($T(\alpha x) = T(x)$ for any $\alpha \neq 0$), and easy
to compute.  Algorithm~\ref{alg:detector} summarizes the procedure.

\begin{algorithm}[H]
\caption{Mimetic adversarial detector.}
\label{alg:detector}
\begin{algorithmic}[1]
\Require image $x \in \mathbb{R}^{H \times W}$, mimetic order $k \in \{2,4,6,8\}$, threshold $\tau$.
\State $G \gets \texttt{grad2D}(k, W, 1, H, 1)$ \Comment{MOLE \cite{MOLE}}
\State $P \gets \texttt{weightsP2D}(k, W, 1, H, 1)$
\State $u \gets$ vectorize $x$ with one replicated ghost layer, $x$-fastest order
\State $E_{H^1} \gets (G\,u)^{\!\top}\, P\,(G\,u)$
\State $E_{\ell^2} \gets \sum_{i,j} x_{ij}^2$
\State $T \gets E_{H^1} \,/\, E_{\ell^2}$
\State \Return \texttt{adversarial} if $T > \tau$, else \texttt{clean}
\end{algorithmic}
\end{algorithm}

The threshold $\tau$ is calibrated once on a held-out clean set as the
$(1-\alpha)$-quantile of $T$ for the desired false-positive rate $\alpha$.
Total cost: $O(HW)$ floating-point operations per image, dominated by a single sparse matrix-vector product.

\section{Test case: peppers}

\paragraph{Setup.}
We attack \texttt{peppers.png} (from MATLAB's Image Processing Toolbox), resized to the
$227 \times 227$ RGB input of a pretrained \textbf{SqueezeNet}
\cite{Iandola2016}.  On the clean image SqueezeNet predicts
\emph{bell pepper} with confidence $0.41$.  We construct four inputs:
\begin{itemize}[leftmargin=*,itemsep=2pt]
   \item \textbf{Clean:} no perturbation.
   \item \textbf{FGSM:} one untargeted fast-gradient-sign step,
         $\delta_\text{FGSM} = \varepsilon \,\operatorname{sign}
         \bigl(\nabla_x \mathcal{L}(f, x, \hat y)\bigr)$, where $\hat y$ is
         the clean prediction.  It flips SqueezeNet to \emph{doormat}.
   \item \textbf{PGD:} $20$-step projected gradient ascent on the same
         loss (no random initialization), step size $2/255$, projected
         after each step onto the $\ell^\infty$ ball of radius
         $\varepsilon$.  It flips SqueezeNet
         to \emph{maraca} with confidence $\approx 1$.
   \item \textbf{Smooth control:}
         $(\delta_\text{low})_{ij} = \varepsilon\,\sin(4\pi i/H)
         \,\sin(4\pi j/W)$, equal $\ell^\infty$ amplitude but no
         high-frequency content.
\end{itemize}
We take $\varepsilon = 16/255$, a standard FGSM/PGD budget at 8-bit
pixel depth, clip pixel values to the valid range after each step, and
form the adversarial images in RGB at the classifier input.  The detector
then converts each input to grayscale and computes $T$; it never sees the
classifier, its gradients, or the attack.  With no random initialization, the FGSM and PGD perturbations
are deterministic given the network, so the values below are exactly
reproducible.

\begin{figure}[H]
\centering
\includegraphics[width=0.95\textwidth]{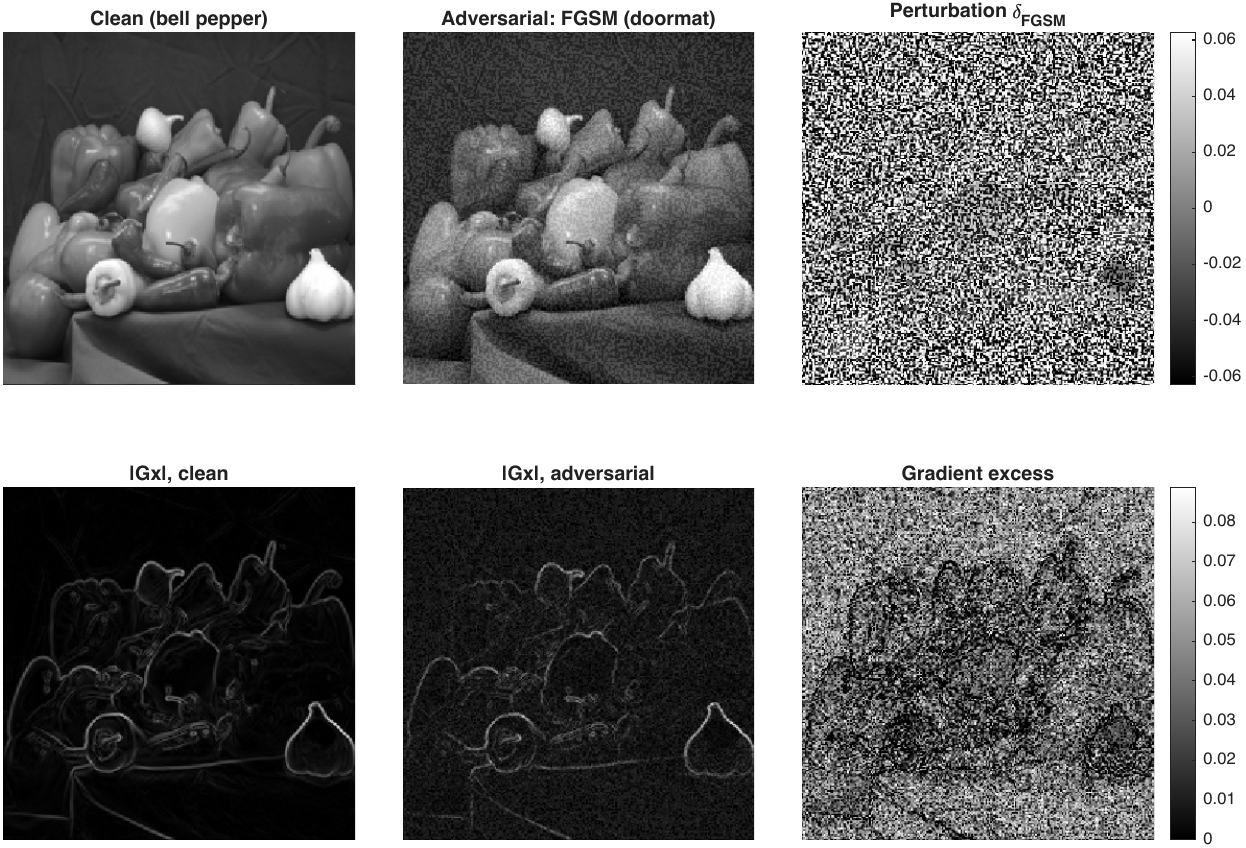}
\caption{Top row: the clean image (SqueezeNet: \emph{bell pepper}), the
FGSM adversarial image $x + \delta_\text{FGSM}$ (SqueezeNet:
\emph{doormat}), and the perturbation $\delta_\text{FGSM}$.  Bottom row:
cell-centered mimetic gradient magnitude $|Gx|$ (order $k=2$) for the clean
and adversarial images, and the gradient excess
$|G(x+\delta_\text{FGSM})| - |Gx|$.  The clean image concentrates its
gradient energy on object edges, whereas the FGSM perturbation leaves a
near-uniform high-frequency imprint across the whole image, exactly what
$T$ measures.}
\label{fig:peppers}
\end{figure}

\paragraph{Results.}
Table~\ref{tab:results} reports $T$ at orders $k = 2, 4, 6, 8$.  At every $k$:
\begin{itemize}[leftmargin=*,itemsep=2pt]
   \item FGSM raises $T$ to $4.1$--$5.0\times$ its clean value; PGD to
         $1.9$--$2.2\times$.  Both sit far above the clean baseline, so a
         single threshold flags either attack.
   \item The smooth control leaves $T$ within $1\%$ of clean
         ($0.99\times$), confirming that the detector responds to the
         high \emph{frequency} of the perturbation, not its amplitude.
\end{itemize}
PGD is the stronger attack at the classifier, driving the prediction
to a wrong class with near-unit confidence, yet it raises $T$ less than
FGSM.  This is expected: FGSM takes the full $\varepsilon$ step on every
pixel and so deposits the maximal high-frequency energy, whereas PGD
reaches misclassification with a more economical, partly sub-maximal
perturbation and is therefore quieter, but still plainly separated from
clean.  The detection margin grows monotonically with $k$ for both
attacks: FGSM from $4.09\times$ at $k=2$ to $5.05\times$ at $k=8$, and PGD
from $1.94\times$ to $2.23\times$.  Higher-order Corbino--Castillo
operators sharpen the discrimination because the high-frequency content of
the adversarial perturbation is mapped more faithfully through their
wider-bandwidth interior stencils. Figure~\ref{fig:peppers} visualizes
this gradient signature.

\begin{table}[H]
\centering
\caption{Mimetic gradient-energy ratio $T$ on the $227 \times 227$
\texttt{peppers} test image at $\varepsilon = 16/255$, for untargeted FGSM
and PGD attacks against SqueezeNet, computed with MOLE's
Corbino--Castillo operators at orders $k \in \{2, 4, 6, 8\}$.
$T_\text{low}$ is the equal-amplitude smooth control.  Ratios are
computed from unrounded values.}
\label{tab:results}
\begin{tabular}{ccccccc}
\toprule
$k$ & $T_\text{clean}$ & $T_\text{FGSM}$ & $T_\text{PGD}$ & $T_\text{low}$
    & $T_\text{FGSM}/T_\text{clean}$ & $T_\text{PGD}/T_\text{clean}$ \\
\midrule
2 & $2.29 \times 10^{-2}$ & $9.37 \times 10^{-2}$ & $4.45 \times 10^{-2}$
  & $2.28 \times 10^{-2}$ & $4.09\times$ & $1.94\times$ \\
4 & $2.52 \times 10^{-2}$ & $1.14 \times 10^{-1}$ & $5.24 \times 10^{-2}$
  & $2.50 \times 10^{-2}$ & $4.54\times$ & $2.08\times$ \\
6 & $2.59 \times 10^{-2}$ & $1.23 \times 10^{-1}$ & $5.57 \times 10^{-2}$
  & $2.57 \times 10^{-2}$ & $4.76\times$ & $2.15\times$ \\
8 & $2.63 \times 10^{-2}$ & $1.33 \times 10^{-1}$ & $5.87 \times 10^{-2}$
  & $2.61 \times 10^{-2}$ & $5.05\times$ & $2.23\times$ \\
\bottomrule
\end{tabular}
\end{table}

\section{Conclusions}

We have presented a single-shot, training-free detector for adversarial
image perturbations, using the Corbino--Castillo mimetic operators from
MOLE.  Three properties make the detector deployable:
\begin{itemize}[leftmargin=*,itemsep=2pt]
   \item It needs no training set of adversarial examples, no surrogate
         classifier, and no access to the network being attacked.
   \item It runs in $O(HW)$ time per image.
   \item Its detection margin sharpens monotonically as the mimetic order
         $k$ is increased, a property that comes for free with
         MOLE's high-order construction.
\end{itemize}
The detector is a \emph{detector}, not a certified defense: an adaptive
attacker who explicitly minimizes $T$ subject to misclassification can
drive the statistic close to its clean value at the cost of a smaller
attack-success rate or a larger $\ell^\infty$ budget.  The FGSM--PGD gap
in Table~\ref{tab:results} is a mild preview of this: PGD already
deposits less gradient energy than FGSM while attacking more successfully,
so an attacker who optimizes against $T$ directly would deposit less
still.  Two natural extensions are to color and multispectral imagery
(per-channel $T$, mixed by a learned linear combination) and to
non-uniform or curvilinear sensor geometries where the mimetic structure
is most powerful.

\end{document}